\documentclass[journal,twoside,web]{ieeecolor}

\usepackage{generic}

\usepackage{amsmath,amsfonts,amssymb}
\usepackage{mathtools}
\usepackage{bbm}
\usepackage{epsfig} %
\usepackage{times} %
\usepackage[dvipsnames]{xcolor}
\usepackage[english]{babel}
\usepackage{graphicx}
\usepackage{subfigure}
\usepackage{nicefrac}
\usepackage{etoolbox}
\usepackage{indentfirst}
\usepackage{caption}
\usepackage{lmodern}
\usepackage{fancyhdr}
\usepackage{algorithm, algorithmic}

\makeatletter
\newcommand\fs@betterruled{%
  \def\@fs@cfont{\bfseries}\let\@fs@capt\floatc@ruled
  \def\@fs@pre{\vspace*{5pt}\hrule height.8pt depth0pt \kern2pt}%
  \def\@fs@post{\kern2pt\hrule\relax}%
  \def\@fs@mid{\kern2pt\hrule\kern2pt}%
  \let\@fs@iftopcapt\iftrue}
\floatstyle{betterruled}
\restylefloat{algorithm}
\makeatother

\definecolor{sztakiblue}{RGB}{0,0,0}

\newcommand{\QE}{Q_{\scriptscriptstyle E}}
\newcommand{\QN}{Q_{\scriptscriptstyle N}}

\newcommand{\BI}{\mathbb{I}}

\newcommand{\CH}{\mathcal{H}}

\def \CD{\mathcal{D}}

\usepackage{mathtools}

\newcommand{\RR}{\mathbb{R}}

\newcommand{\NN}{\mathbb{N}}
\newcommand{\ZZ}{\mathbb{Z}}
\newcommand{\EE}{\mathbb{E}}
\newcommand{\PP}{\mathbb{P}}

\usepackage{calc}

\usepackage{accents}
\newcommand{\dbtilde}[1]{\accentset{\approx}{#1}}
\DeclareFontFamily{U}{mathx}{\hyphenchar\font45}
\DeclareFontShape{U}{mathx}{m}{n}{<-> mathx10}{}
\DeclareSymbolFont{mathx}{U}{mathx}{m}{n}
\DeclareMathAccent{\widebar}{0}{mathx}{"73}

\newtheorem{innercustomgeneric}{\customgenericname}
\providecommand{\customgenericname}{}
\newcommand{\newcustomtheorem}[2]{%
	\newenvironment{#1}[1]
	{%
		\renewcommand\customgenericname{#2}%
		\renewcommand\theinnercustomgeneric{##1}%
		\innercustomgeneric
	}
	{\endinnercustomgeneric}
}

\newcustomtheorem{customthm}{Theorem}
\newcustomtheorem{customlemma}{Lemma}     %

\newtheorem{theorem}{Theorem}
\newtheorem{lemma}{Lemma}
\newtheorem{definition}{Definition}

\newcommand{\norm}[1]{\left\lVert#1\right\rVert}

\usepackage{epsfig} %

\title{\LARGE \bf
Robust Independence Tests with Finite Sample Guarantees\\ 
for Synchronous Stochastic Linear Systems 
}

\author{Ambrus Tam{\'a}s$^{1,2}$ \and\qquad D{\'a}niel {\'A}goston B{\'a}lint$^{3}$ \and\qquad Bal{\'a}zs Csan{\'a}d Cs{\'a}ji$^{1,2}$, \IEEEmembership{Member, IEEE} %
\thanks{*This research was supported by the European Union project RRF-2.3.1-21-2022-00004 within the framework of the Artificial Intelligence National Laboratory. The work was also supported by the TKP2021-NKTA-01 grant of the National Research, Development and Innovation Office (NRDIO), Hungary. The authors acknowledge the professional support of the 
the Cooperative Doctoral Program of the Ministry of Innovation and Technology, as well, financed from an NRDIO Fund.}%
\thanks{$^{1}$A.~Tam\'as and B.~Cs.~Cs\'aji are with SZTAKI: Institute for Computer Science and Control, E\"otv\"os Lor\'and Research Network, Budapest, Hungary,\, {\tt\small tamasamb@sztaki.hu},\, {\tt\small csaji@sztaki.hu}}%
\thanks{$^{2}$A.~Tam\'as and B.~Cs.~Cs\'aji are also with Institute of Mathematics, E\"otv\"os Lor\'and University (ELTE), Budapest, Hungary}%
\thanks{$^{3}$D.~{\'A}.~B{\'a}lint is independent researcher, financial support from Swissblock Technologies for the second half of the project is gratefully acknowledged, ~{\tt\small bdagoston@gmail.com}}%
}

\begin{document}

\newcounter{asscounter}
\setcounter{asscounter}{1}

\maketitle
\thispagestyle{empty}
\pagestyle{empty}

\begin{abstract}

The paper introduces robust independence tests with non-asymptotically guaranteed significance levels for 
stochastic linear time-invariant systems, assuming that the observed outputs are 
synchronous,
which means that the systems are driven by jointly i.i.d.\ noises. Our method provides bounds for the type I error probabilities that are distribution-free, i.e.,
the innovations can have arbitrary distributions.
The algorithm combines confidence region estimates with permutation tests and general dependence measures, such as the Hilbert--Schmidt independence criterion and the distance covariance, to detect any nonlinear dependence between the 
observed systems. We also prove the consistency of our hypothesis tests under mild assumptions and 
demonstrate the ideas through the example of 
autoregressive systems.

\end{abstract}

\begin{IEEEkeywords}
independence tests, stochastic linear systems, distribution-free guarantees, dependence measures
\end{IEEEkeywords}

\section{Introduction}
\IEEEPARstart{S}{tatistical} independence is a key notion in several areas of statistics and probability theory, including system identification \cite{ljung1999system}, time-series analysis, signal processing and machine learning. %
In this paper we present a non-asymptotic framework to construct hypothesis tests for the independence of two simultaneous linear systems or time series. Our setup is distribution-free, i.e., the process noises can follow any probability law. The presented hypothesis tests might be used for instance to identify (conditionally) dependent price returns in the stock market or to find interconnections between two systems in a network.
Other potential applications include biological systems, where the independence of cell mechanisms may be tested, or one could
analyze whether social phenomena which occur simultaneously are dependent. 

Given an i.i.d.\ sample of random pairs with some joint distribution, there already exists several hypothesis tests for independence, e.g., the celebrated $\chi^2$ test, %
Hoeffding's test based on the factorization of the joint distribution function,
and Hilbert--Schmidt independence criterion (HSIC) based tests \cite{gretton2007kernel}.
These methods typically use the limiting distribution of some test statistic to calculate the $p$-values
for a given sample size. %
Usually it is challenging or even impossible to calculate the exact distribution of these 
test
statistics, but 
permutation and Monte Carlo tests offer viable options.
For time series, 
it is even more challenging to test independence. For this, 
HSIC \cite{chwialkowski2014kernel} and distance correlation \cite{szekely2009brownian} based approaches were proposed, which are only supported by asymptotic guarantees.
\section{Problem Setting}

We construct robust permutation-based independence tests \cite{pfister2018kernel,rindt2021consistency}, for linear systems in a general setting \cite{ljung1999system}.
Let us consider two scalar (discrete-time, time-invariant) stochastic linear systems with general dynamics:
\begin{equation}
\label{eq:twosys}
\begin{aligned}
    Y_t = &\; G_1(q^{-1}; \theta_*) U_t + H_1(q^{-1}; \theta_*) E_t,\\[1mm]
    Z_t = &\; G_2(q^{-1}; \gamma_*) V_t + H_2(q^{-1}; \gamma_*) N_t, 
\end{aligned}
\end{equation}
 for $t \in \ZZ$, where $U_t$ and $V_t$ are (exogenous) inputs; $Y_t$ and $Z_t$ are (observable) outputs; $q^{-1}$ is the backshift operator given by $q^{-1} X_t \doteq X_{t-1}$ for any time series $\{X_t\}$; 
 $E_t$ and $N_t$ are possibly dependent process noises; $G_1, H_1$, and $G_2, H_2$ are rational transfer functions determined by finite dimensional parameter spaces $\Theta$ and $\Gamma$, respectively. The unknown true 
 parameters are $\theta_*$ and $\gamma_*$. 

 Our main assumptions are as follows, cf. \cite{ljung1999system}:
 \smallskip
\begin{itemize}
        \item[A\theasscounter] {\em The true systems generating outputs $\{ Y_t\}$ and $\{Z_t\}$ are in the model classes, i.e., $\theta_* \in \Theta$ and $\gamma_* \in \Gamma$.}
        \addtocounter{asscounter}{1}
        \item[A\theasscounter] {\em Rational (causal) transfer functions $G_1$, $G_2$, $H_1$ and $H_2$ have known orders.}
        \addtocounter{asscounter}{1}
        \item[A\theasscounter] {\em $H_1$ and $H_2$ are invertible for all $\theta \in \Theta$ and $\gamma \in \Gamma$.}
        \addtocounter{asscounter}{1}
        \item[A\theasscounter] {\em The systems are initialized in zero, i.e., $Y_t = Z_t = U_t = V_t = E_t = N_t=  0$, for all $t \leq 0$.} \addtocounter{asscounter}{1}
        \item[A\theasscounter] {\em The systems are driven by an {\em i.i.d.} innovation sequence $\{(E_t,N_t)\}_{t = 1}^{\infty}$ from the distribution of $(E, N)$. } \addtocounter{asscounter}{1}
        \item[A\theasscounter] {\em The systems operate in open-loop: the inputs $\{U_t\}$, $\{ V_t\}$ are independent of the noises $\{E_t\}$, $\{ N_t\}$.}\addtocounter{asscounter}{1}
\end{itemize}
\smallskip 
\noindent
ARMAX models, e.g., can satisfy these conditions. For simplicity, we treat the inputs $\{U_t\}$, $\{ V_t\}$ as deterministic sequences. This is w.l.o.g.\ as we can always condition on the inputs, as they are independent of the innovations. In case the inputs are stochastic, the obtained results should be interpreted as conditional to the inputs, i.e., we test whether $\{Y_t\}$ and $\{Z_t\}$ are conditionally independent given the inputs $\{U_t\}$, $\{ V_t\}$. Also, we can assume that the parameterization is unique, e.g., by assuming w.l.o.g.\ that $H_1(0; \theta)=1$ and $H_2(0; \gamma) = 1$ for all $\theta  \in \Theta, \gamma \in \Gamma$.
 
In this paper we aim at constructing consistent hypothesis tests with finite sample guarantees for the independence of output sequences $\{Y_t\}$ and $\{Z_t\}$. For this we observe that if $\{Y_t\}$ and $\{Z_t\}$ are driven by a jointly i.i.d. noise $(E,N)$, see A5, with joint distribution $Q_{\scriptscriptstyle{E,N}}$ and marginals $\QE$ respectively $\QN$, then the independence of $\{E_t\}$ and $\{N_t\}$ is equivalent to the independence of $\{Y_t\}$ and $\{Z_t\}$ conditional on the inputs. Therefore, it is sufficient to test the null hypothesis 
\vspace{-0.5mm}
\begin{equation}
    H_0: Q_{\scriptscriptstyle{E,N}} = \QE \otimes \QN\qquad H_1: Q_{\scriptscriptstyle{E,N}} \not = \QE \otimes \QN
\vspace{-0.5mm}    
\end{equation}
The main challenge is that the parameters $\theta^*, \gamma^*$ are unknown, henceforth the noise terms are not observable.

For simplicity, we assume that the finite sample of inputs, $\{U_t\}$, $\{V_t\}$, and outputs, $\{Y_t\}$, $\{Z_t\}$, available for estimation is large enough to compute $n$ prediction errors $\{(E_t(\theta), N_t(\gamma))\}_{t=1}^n$ for any values of $\theta$ and $\gamma$.

We construct the hypothesis tests in several steps. First we estimate the system parameters with non-asymptotic confidence regions, then we reconstruct the residuals on the set of possible parameters and apply permutation tests on them. Our main assumption is that the linear systems are driven simultaneously, see assumption $A5$, and that the noise terms could be recovered if the system parameters were known, see assumption $A3$. We only rely on the i.i.d.\ assumption to quantify the user--chosen probability of type~I error and prove that the probability of type~II error vanishes asymptotically. 

\section{Permutation Tests for the I.I.D.\ Case}

First, for simplicity, assume that the parameters are known. In this case we can compute the noise terms as 
\begin{equation}
\begin{aligned}
    E_t\, & =\, E_t(\theta^*) =\, H^{-1}_1(q^{-1}; \theta_*)( Y_t - G_1(q^{-1};\theta_*) U_t), \\[0mm]
    N_t\, & =\, N_t(\gamma_*) =\, H^{-1}_2(q^{-1}; \gamma_*)( Z_t - G_1(q^{-1};\gamma_*) V_t),
\end{aligned}
\end{equation}
using $A3,A4$ and $A6$. Then the independence of $E$ and \hspace{1mm} $N$ can be tested based on the i.i.d.\ sample, $\{(E_t, N_t)\}_{t=1}^n$.

\subsection{Resampling}

Let $\mathcal{D}_0 \doteq \{(E_t,N_t)\}_{t=1}^n$ be the known noise terms. We propose a rank test which is based on empirical dependence measure values calculated from perturbed samples. The idea is to generate new, perturbed datasets which have the same distributional properties to the original observations in case the null hypothesis is true.

We apply the permutation test which was first presented in \cite{pfister2018kernel}; proof of consistency and further ramifications have been provided in \cite{rindt2021consistency}.
We choose an arbitrary (rational) significance level $\alpha$ in advance and %
integer hyperparameters $1 \leq r \leq p \leq m$ such that
\begin{equation}
    \alpha = 1 - \frac{p- r + 1}{m}.
\end{equation}
Let $S_n$ be the set of permutations on $[n]\doteq \{1, \dots , n\}$ and $\{\pi_{j}\}_{j=1}^{m-1}$ be uniformly randomly generated from $S_n$. We construct $m-1$ new alternative samples by
\begin{equation}
    \CD_j = \pi_j \CD_0 \doteq\{ (E_i, N_{\pi_j(i)})\}_{i=1}^n
\end{equation}
for $j= 1, \dots, m-1$. Altogether we end up having $m$ datasets to compare. Observe that if $H_0$ holds true then $\CD_0 = \big((E_1, N_1),\dots, (E_n,N_n)\big)$ has the same distribution as $\CD_j = \big((E_1, N_{\pi_j(1)}),\dots, (E_n,N_{\pi_j(n)})\big)$ for any permutation $\pi_j$, whereas if $H_1$ holds then the distribution of $\CD_j$ is different from that of $\CD_0$. Our goal is to quantify this difference whenever the null hypothesis does not hold.

\subsection{Exact Coverage}

The comparison of the datasets is carried out with the help of ranking functions, \cite{vovk2005algorithmic, csajitamas2019}. Let $\psi$ be a ranking function that orders the datasets in a total order, i.e.,
\smallskip
{\em 
\begin{definition}
Let $A$ be a set. We say that $\psi: A^m \to [m]$ is a ranking function if it has the two properties below:
\begin{enumerate}
    \item For all $a_1, \dots, a_m \in A$ and for all permutation $\tau: \{2, \dots , m\} \to \{2, \dots, m\}$ we have that
    \vspace{-1mm}
    \begin{equation}
        \psi(a_1, \dots, a_m ) = \psi(a_1, a_{\tau(2)}, \dots, a_{\tau(m)}).
    \vspace{-1mm}        
    \end{equation}
\item If $a_i \neq a_j$, then $\psi(a_i, \{a_k\}_{k\neq i}) \neq \psi(a_j, \{a_k\}_{k\neq j})$.
\end{enumerate}
\end{definition}
}
\medskip

Our main tool for quantifying the probability of type~I error will be the following lemma, see \cite[Lemma 1]{csajitamas2019}:
\smallskip
\begin{lemma}\label{lemma:uniform}
{\em Let $\xi_1, \dots, \xi_m$ be a.s.\ pairwise different exchangeable variables and let $\psi$ be a ranking function. Then $\psi(\xi_1, \dots, \xi_m)$ has a uniform distribution on $[m]$}.
\end{lemma}
\smallskip
A technical challenge is posed by identical datasets. We use a random permutation $\sigma$ on $[m-1]_{\scriptscriptstyle 0} \doteq \{0, \dots, m-1\}$ independently generated from everything else to resolve this issue, i.e., let $\CD_j^\sigma \doteq (\CD_j, \sigma(j))$ for $j \in[m-1]_{\scriptscriptstyle 0}$.

\smallskip
{\em
\begin{theorem}\label{thm:exact_1}
Assume $A1$ -- $A6$ and that $\theta_*$, $\gamma_*$ and $\psi$ are given. If $H_0$ holds true then for any $1 \leq r \leq p \leq m$:
\vspace{-0.5mm}
\begin{equation}
\PP\,\Big(\,r\leq \psi(\CD_0^\sigma, \dots, \CD_{m-1}^\sigma) \leq p \,\Big)\, =\, \frac{p-r+1}{m}.
\end{equation}
\end{theorem}
}
\medskip
\begin{proof}
     Notice that $\{\CD_j^\sigma\}_{j=0}^{m-1}$ are almost surely pairwise different exchangeable variables under $H_0$, therefore Theorem \ref{thm:exact_1} follows from Lemma \ref{lemma:uniform}.
\end{proof}
Observe that Theorem \ref{thm:exact_1} is completely distribution-free and provides us finite sample guarantees for the type~I error probabilities. In addition, the significance level of the proposed scheme is exact and user-chosen (rational).
In the following section we construct ranking functions that ensure the consistency of the proposed test.

\subsection{Dependence Measures}

Dependence measures are used for assessing dependency between random variables $E$, $N$ with joint distribution $Q_{\scriptscriptstyle{E,N}}$, or, in the empirical case, i.i.d. datasets $\CD_0 = \{(E_i, N_i)\}_{i=1}^n$ generated from $Q_{\scriptscriptstyle{E,N}}$. A dependence measure, $\Delta$, has two properties. First, it needs to be {\em characteristic}, i.e., $\Delta(E,N) = 0$ if and only if $E$ and $N$ are independent. Second, it needs to exhibit a {\em consistent estimator} $\widehat{\Delta}$, that is (as the sample size $n$ increases)
\begin{equation}\label{eq:8}
    \widehat{\Delta}(\CD_0) \,\doteq\, \widehat{\Delta}(\{(E_i, N_i)\}_{i=1}^n)\xlongrightarrow{p} \Delta(E, N).
\end{equation}
We consider the following two dependence measures.

\subsubsection{Hilbert--Schmidt Independence Criterion}

Let $\CH_k$ be a reproducing kernel Hilbert space (RKHS) and $k:\RR \times \RR \to \RR$ be its reproducing kernel. 
For a random variable $E$ with $\EE \sqrt{k(E,E)} < \infty$, the distribution $\QE$ of $E$ can be embedded into $\CH_k$ by
    $\mu(\QE) \doteq \EE[ k(E, \cdot) ],$
where the expectation is a Bochner integral; $\mu(\QE)$ is called the {\em kernel mean embedding} of $\QE$ \cite{smola2007hilbert}. 
Kernel $k$ is called {\em characteristic} if $\mu$ is injective.

In what follows, let $k$ and $\ell$ be positive definite kernels. It is well-known that the (tensor) product kernel
\begin{equation}
\begin{aligned}
    &k\otimes \ell: (\RR \times \RR)^2 \to \RR\\
    & \hspace{1cm}\big((x_1, x_2),(y_1, y_2)\big) \mapsto k(x_1, x_2)\cdot \ell(y_1,y_2)
\end{aligned}
\end{equation}
is also positive definite. With this object, the (centered) cross--covariance operator is defined as
\begin{equation*}
\begin{aligned}
    \mathbf{C}_{\scriptscriptstyle{E,N}} &\doteq \EE \big[ k\otimes \ell \big((E,N), (\cdot, \cdot )\big)\big] - \EE [k(E, \cdot )] \otimes \EE[\ell(N, \cdot)]\\[1mm]
    & = \mu( Q_{\scriptscriptstyle{E,N}}) - \mu( \QE\otimes\QN).
\end{aligned}  
\end{equation*}
Note that one can recover the standard covariance by applying linear kernels, $k(x, y) = \ell(x, y) = x\cdot y$.

With this notation, we are now able to recall the definition of Hilbert--Schmidt independence criterion \cite{gretton2007kernel}:
\begin{equation}
    \text{HSIC}(Q_{\scriptscriptstyle{E,N}}, \CH_k, \CH_l) \doteq \norm{\mathbf{C}_{\scriptscriptstyle{E,N}}}_\otimes^2,
\end{equation}
where $\|\cdot \|_\otimes$ denotes the norm of the product RKHS. If $k$, $\ell$ and $Q_{\scriptscriptstyle{E,N}}$ are fixed, we may write $\text{HSIC}(E,N) \doteq \text{HSIC}(Q_{\scriptscriptstyle{E,N}}, \CH_k, \CH_l)$ and use a more intuitive form:
\begin{equation}
\hspace*{-2mm}
\label{hsic_intuitive}
\begin{aligned}
    &\text{HSIC}(E,N) = \EE[ k(E,E') \ell(N,N')]\\[1mm]
    & + \EE[ k(E,E')] \EE[ \ell(N,N')]- 2 \EE[ k(E,E') \ell(N,N'')],
\end{aligned}
\end{equation}
where $(E,N)$, $(E', N')$ and $(E'',N'')$ are i.i.d. copies from $Q_{\scriptscriptstyle{E,N}}$. %
HSIC is characteristic if $k\otimes \ell$ is characteristic.

Formula \eqref{hsic_intuitive} motivates the empirical estimate:
\vspace{-1mm}
\begin{equation}\label{eq:hsic}
\begin{aligned}
    &\text{HSIC}_n(\CD_0) \doteq \frac{1}{n^2} \sum_{(i,j) \in [n]^2} k(E_i,E_j)\ell(N_i,N_j)\\
     & \hspace*{-1mm}+\! \frac{1}{n^4} \hspace*{-5.8mm}\sum_{\hspace*{2mm}(i,j,r,s) \in [n]^4} \hspace*{-6.3mm} k(E_i,E_j) \ell(N_r,N_s) \!- \!\frac{1}{n^3} \hspace*{-5.3mm}\sum_{\hspace*{2mm}(i,j,s) \in [n]^3} \hspace*{-4.8mm}k(E_i,E_j)\ell(N_i,N_s).
\end{aligned}
\vspace{-1mm}
\end{equation}
The consistency of $\text{HSIC}_n$ is proved in \cite{pfister2018kernel}. 

\subsubsection{Distance Covariance}

Distance covariance was first introduced in \cite[Definition 2.]{szekely2007measuring} and the definition below is due to \cite[Theorem 8.]{szekely2009brownian}. For $E$, $N$ with finite expectations, the distance covariance is defined as
\begin{equation*}
\begin{aligned}
    & \text{dCov}^2(E,N)  \doteq \EE\norm{E-E'}\norm{N-N'} \\
    & + \EE\norm{E-E'}\EE\norm{N-N'} - 2\EE\norm{E-E'}\norm{N-N''},
\end{aligned}
\end{equation*}
where $(E,N)$, $(E', N')$ and $(E'',N'')$ are i.i.d.\ copies from $Q_{\scriptscriptstyle{E,N}}$. Distance covariance has some excellent properties in terms of measuring independence, most importantly, it is characteristic \cite[Theorem 3.]{szekely2007measuring}.

The doubly centered distances are $A_{j, k} = a_{j, k} - a_{j\cdot} - a_{\cdot k} + a_{\cdot\cdot}$ where $a_{j, k } = \norm{E_j - E_k}$, $a_{j\cdot} = \sum_{k=1}^n a_{j, k}/n$, $a_{\cdot k} = \sum_{j=1}^n a_{j, k}/n$ and $a_{\cdot\cdot} = \sum_{j, k=1}^n a_{j, k}/(n^2)$; and let $B$ be analogously for $\{N_i\}_{i=1}^n$. Then, the empirical distance covariance \cite[Definition 4.]{szekely2007measuring}
can be computed as:
\begin{equation}
\begin{aligned}
    \text{dCov}^2_n (\CD_0) & \doteq \frac{1}{n^2} \sum_{j=1}^n \sum_{k=1}^n A_{j, k} B_{j, k}.
\end{aligned}
\end{equation}
This empirical estimate is consistent \cite[Theorem 2]{szekely2007measuring}.

\vspace*{-1mm}
\subsection{Hypothesis Test for Independence}

Our hypothesis test compares empirical dependence measure estimates via a ranking function $\psi$. To resolve (the unlikely event of) ties during comparison, we amend the order function $\prec_\sigma$;
for technical details, see \cite{csaji2014sign}. Let
\vspace{-1mm}
\begin{equation}
\begin{aligned}
    &\widehat{\Delta}_{n}^{(j)} = |\widehat{\Delta}(\CD_j)|\quad\quad \text{for }j=0,1, \dots, m-1\; \text{ and}\\[0.5mm]
    &\psi_\Delta(\CD_0^\sigma, \dots , \CD_{m-1}^\sigma) \doteq 1 + \sum_{j=1}^{m-1} \BI\,\Big(\,\widehat{\Delta}_n^{(0)} \prec_\sigma \widehat{\Delta}_n^{(j)}\,\Big).
\end{aligned}
\vspace{-0.5mm}
\end{equation}
We choose an integer $r$ for significance level $\alpha$ in such a way that $\alpha = r\,/\,m$ and reject $H_0$ if and only if $\psi_\Delta( \CD_0^\sigma, \dots, \CD_{m-1}^\sigma) \leq r$. 
 If $H_0$ holds then $\{\CD_j^\sigma\}_{j=0}^{m-1}$ are exchangeable, hence the rank statistic is uniform. If $H_0$ does not hold, then $\Delta(E,N) \neq 0$ and by \eqref{eq:8}, variable $\widehat{\Delta}(\CD_0)$ tends to a positive number. However, the perturbed samples $\{\CD_j^\sigma\}_{j=1}^{m-1}$ are almost i.i.d.\,, because the pairs are shuffled, thus one expects that $\widehat{\Delta}(\CD_j)$ tends to $0$ for $j\in [m-1]$. In conclusion asymptotically $\widehat{\Delta}(\CD_0)$ dominates $\widehat{\Delta}( \CD_j)$ for every $j \in[m-1]$.
The hypothesis test is summarised in Algorithm 1.
\setlength{\textfloatsep}{1mm}
\begin{algorithm}[t]
    \label{hypothesis_algo}
    \caption{Independence Test for an I.I.D.\ Sample}
    \textbf{Inputs:} i.i.d. sample $\CD_0$, desired significance level $\alpha$,\\[0.5mm]
         \hspace*{12.6mm} dependence measure estimator $\widehat{\Delta}$\\[0.5mm]
         \hspace*{12.6mm} tie breaking permutation $\sigma$ on $[m-1]_{\scriptscriptstyle 0}$\\[-2mm]
    \begin{algorithmic}[1]
         \hrule
         \STATE Choose integers $1 \leq r \leq m$ such that $\alpha = \nicefrac{r}{m}$.\\[1mm]
         \STATE Generate $m-1$ random permutation $\{\pi_j\}_{j=1}^{m-1}$ \\[1mm]
         uniformly from $S_n$.\\[1mm]
         \STATE Construct $m-1$ new alternative sample \\[1mm]
         by $\CD_j= \{(E_i,N_{\pi_j(i)})\}_{i=1}^n$ for $j \in [m-1]$\\[1mm]
         and let $\CD_j^\sigma \doteq (\CD_j, \sigma(j))$ for $j \in [m-1]_{\scriptscriptstyle 0}$.
         \STATE Calculate the dependence measure estimates \\[1mm]
         $\widehat{\Delta}_n^{(j)} = |\widehat{\Delta}(\CD_j)|$ for $j = 0, 1, \dots, m-1$.\\[1mm]
         \STATE Compute the rank statistic:\vspace*{-2mm}
         \[\psi_\Delta( \CD_0^\sigma, \dots, \CD_{m-1}^\sigma)= 1 + \sum_{j=1}^{m-1} \BI\,\big(\,\widehat{\Delta}_n^{(0)} \prec_\sigma \widehat{\Delta}_n^{(j)}\,\big).\vspace*{-2mm}\]
         \STATE Reject $H_0$ if and only if \[\psi_\Delta( \CD_0^\sigma, \dots, \CD_{m-1}^\sigma) \leq r.\]
    \end{algorithmic}
\end{algorithm}
Our exact coverage result below is a direct consequence of  Theorem \ref{thm:exact_1}:
\smallskip
{\em
\begin{theorem}\label{cor:cons}
Assume $A1$ -- $A6$. Let $\widehat{\Delta}$ be a dependence measure estimator, $\psi_\Delta$ be the corresponding ranking function and $r \leq m$ be integers. If $H_0$ holds, then
\begin{equation}
    \PP\,\Big(\psi_\Delta(\CD_0^\sigma, \dots, \CD_{m-1}^\sigma) \leq r \,\Big)\, =\, \frac{r}{m}.
    \vspace*{1.5mm}
\end{equation}
\end{theorem}
}

\subsection{Strong Consistency}

In this section we give asymptotic bounds for the type~II error probability. Assume:
\smallskip
\begin{itemize}
    \item[A\theasscounter] {\em A characteristic dependence measure $\Delta$ and a consistent estimator is given such that for $j \in [m-1]$
        \begin{equation*}
        \begin{aligned}
            \widehat{\Delta}_n^{(0)} \xlongrightarrow{p} |\Delta(E,N)| \qquad \text{and}\qquad \widehat{\Delta}_n^{(j)} \xlongrightarrow{p}  0.
        \end{aligned}
        \end{equation*}
        \addtocounter{asscounter}{1}
        }
\end{itemize}

Assumption $A7$ ensures that the empirical estimates of the used dependence measure vanish for the permuted samples. This assumption is satisfied, for example, by the HSIC-based ranking, see \cite[Lemma 1]{rindt2021consistency}.
{\em
\begin{theorem}\label{thm:consistency_1}
    Assume that $A1$ -- $A7$ hold and $\theta_*$, $\gamma_*$ are given. If $H_1$ holds, then for any $r \geq 1:$
    \begin{equation}\label{eq:prob-rank}
        \PP\big(\, \psi_\Delta(\CD_0^\sigma, \dots, \CD_{m-1}^\sigma) \leq r \,\big) \xrightarrow{\,n\to \infty\,} 1.
    \end{equation}
\end{theorem}
}
\smallskip\vspace*{1mm}
\begin{proof}
    One can bound the probability in \eqref{eq:prob-rank} as
    \begin{equation*}
    \begin{aligned}
         &\PP (\psi_\Delta \leq r) \geq \PP(\psi_\Delta \leq 1) \geq \PP\big( \forall j \in [m-1]: \widehat{\Delta}_n^{(0)} > \widehat{\Delta}_n^{(j)}\big)\\
         &= 1- \PP\big( \,\exists j \in [m-1]:\widehat{\Delta}_n^{(0)} \leq \widehat{\Delta}_n^{(j)}\,\big)\\
         &\geq 1- \sum_{j=1}^{m-1} \PP\big(\, \widehat{\Delta}_n^{(0)} \leq \widehat{\Delta}_n^{(j)}\,\big) \geq 1- m\cdot\PP\big( \,\widehat{\Delta}_n^{(0)} \leq \widehat{\Delta}_n^{(1)}\,\big),
    \end{aligned}  
    \vspace{-0.5mm}
    \end{equation*}
    where $\PP\big( \,\widehat{\Delta}_n^{(0)} \leq \widehat{\Delta}_n^{(1)}\,\big)$ goes to zero, because of $A7$.
\end{proof}
By Theorem \ref{thm:consistency_1} the probability of type~II error tends to zero as the sample size goes to infinity, assuming that the applied dependence measure is characteristic.

\section{Robust Independence Tests}
We now turn our attention to the general problem when the true parameters are unknown constants. In this case, the exact noise terms cannot be computed, but only estimated. 
Assume we can build non-asymptotically guaranteed confidence sets for the true parameters. The idea is then to use a two-step algorithm: first, we construct these (distribution-free) confidence regions, and then apply a parameter-dependent version of the above hypothesis test on each parameter in the confidence set.

\subsection{Parameter-Dependent Hypothesis Test}

We present a meta-algorithm that can work with any confidence region construction for $\theta_*$ and $\gamma_*$. We assume that we have confidence sets $\widehat{\Theta}_n$ and $\widehat{\Gamma}_n$ such that
\begin{equation}
    \label{eq:conf-sets}
    \PP\big( \, \theta_* \in \widehat{\Theta}_n \,\big) \geq 1- \beta\quad\text{and}\quad \PP\big(\, \gamma_* \in \widehat{\Gamma}_n\, \big) \geq 1 - \beta
\end{equation}
hold for all sample size $n \in \NN$ and for some significance level $\beta \in (0,1)$. For simplicity, we will omit $n$ from the notation. By $A3$ we can obtain $E_t(\theta)$ and $N_t(\theta)$ for any parameter-pair candidate $(\theta, \gamma) \in \widehat{\Theta} \times \widehat{\Gamma}$ by 
\begin{equation}
\begin{aligned}
    E_t(\theta)& \doteq H_1^{-1} (q^{-1}, \theta)\big( Y_t -  G_1(q^{-1}, \theta) U_t \big),\\[1mm]
    N_t(\gamma)& \doteq H_2^{-1} (q^{-1}, \gamma)\big( Z_t -  G_2(q^{-1}, \gamma) V_t \big),
\end{aligned}
\end{equation}
for $t = 1, \dots, n$. These quantities can be perturbed as before to construct parameterized alternative datasets
\begin{equation}
\begin{aligned}
    \CD_{j}(\theta,\gamma)= \{(E_i(\theta),N_{\pi_j(i)}(\gamma ))\}_{i=1}^n
\end{aligned}
\end{equation}
for $j=1,\dots, m-1$ and extended by $\sigma$ as before to create $\CD_{j}^\sigma(\theta,\gamma)$ for $j=0, \dots, m-1$. Finally, for any ranking function $\psi$ one can define parameter-dependent ranks as
\begin{equation}
\begin{aligned}
    \psi(\theta, \gamma) \doteq \psi(\CD_0^\sigma(\theta, \gamma), \dots, \CD_{m-1}^\sigma(\theta, \gamma)).
\end{aligned}
\end{equation}

{\em
\begin{theorem}\label{thm:exact}
Assume that $A1$ -- $A6$ hold. Let $\psi$ be any ranking function, $\widehat{\Theta}$ and $\widehat{\Gamma}$ conservative confidence sets with significance level at most $\beta$. If $H_0$ holds true then
\vspace{-0.5mm}
\begin{equation}\label{eq:cons-bound}
\PP\,\Big(\,\max_{(\theta, \gamma) \in \widehat{\Theta} \times \widehat{\Gamma}} \psi(\theta, \gamma) \leq r \,\Big)\, \leq\, \frac{r}{m} + 2 \beta.
\vspace{2mm}
\end{equation}
\end{theorem}
}
\begin{proof}
Using the union bound and \eqref{eq:conf-sets},
one can show that $\widehat{\Theta} \times \widehat{\Gamma}$ is a 
confidence region for $(\theta_*, \gamma_*)$, i.e.,
\begin{equation*}
\begin{aligned}
    \PP\big( \, (\theta_*,\gamma_*) \notin \widehat{\Theta} \times \widehat{\Gamma} \,\big)  &\leq \PP\big(\,\theta_* \notin \widehat{\Theta}\,\big) + \PP\big(\,\gamma_* \notin \widehat{\Gamma}\,\big) \leq 2\beta.
\end{aligned}
\end{equation*}
Then, under $H_0$, we have
\vspace{-1mm}
\begin{equation*}
\begin{aligned}
    &\PP\Big(\,\max_{\theta \in \widehat{\Theta}, \gamma \in \widehat{\Gamma}}
    \psi(\theta, \gamma) 
    \leq r \,\Big)\\
    & = \PP \Big(\,\{(\theta_*,\gamma_*) \notin \widehat{\Theta} \times \widehat{\Gamma}\} \cap \{\max_{\theta \in \widehat{\Theta}, \gamma \in \widehat{\Gamma}} \psi(\theta, \gamma) \leq r \}\,\Big)\\
    &+ \PP \Big(\,\{(\theta_*,\gamma_*) \in \widehat{\Theta} \times \widehat{\Gamma}\} \cap \{\max_{\theta \in \widehat{\Theta}, \gamma \in \widehat{\Gamma}} \psi(\theta, \gamma) \leq r \}\,\Big)
    \\ 
    &\leq \PP \big(\,(\theta_*,\gamma_*) \notin \widehat{\Theta} \times \widehat{\Gamma}\, \big) + \PP ( \,\psi(\theta_*, \gamma_*) \leq r\,),
\end{aligned}
\end{equation*}
where the first term is less than $2\beta$ because of \eqref{eq:cons-bound} and the second term is $r/m$ because of Theorem \ref{thm:exact_1}.
\end{proof}

\subsection{Dependence Measure Ranking}

Ranking functions can be defined similarly to the i.i.d.\ case via dependence measures. The idea is to compute dependence measure estimates w.r.t.\ plausible parameters. Let $\Delta$ be some characteristic dependence measure and $\widehat{\Delta}$ be its (consistent) estimator as above.
Let us define the dependence measure estimate functions as
\begin{equation}
    \widehat{\Delta}_n^{(j)}(\theta, \gamma) \,\doteq\, |\widehat{\Delta}(\CD_j(\theta, \gamma))|,
\end{equation}
for $j = 0, 1, \dots, m-1$ and the ranking function as
\begin{equation}
\begin{aligned}    
    \psi_\Delta(\theta, \gamma) \,\doteq\, 1 + \sum_{j=1}^{m-1} \BI\,\big(\,\widehat{\Delta}_n^{(0)}(\theta, \gamma) \prec_\sigma \widehat{\Delta}_n^{(j)}(\theta, \gamma)\,\big).
\end{aligned}
\vspace{-2mm}
\end{equation}
If $H_0$ does not hold, then $\widehat{\Delta}_n^{(0)}(\theta, \gamma)$ tends to be the greatest around $(\theta_*, \gamma_*)$, henceforth, we reject the null hypothesis if $\psi_\Delta(\theta, \gamma)$ is at most some user-chosen rank value $r$ on $\widehat{\Theta} \times \widehat{\Gamma}$. The step--by--step method is presented in the pseudocode of Algorithm 2. Note that the test exhibits finite sample guarantees for the significance level, as it is showed by Theorem~\ref{thm:exact}.

\setlength{\textfloatsep}{3mm}
\begin{algorithm}[t]
    \label{alg_2}
    \caption{\\Independence Test for Synchronous Systems}
         \textbf{Inputs:} observations $\{U_t\}$, $\{Y_t\}$, $\{V_t\}$ and $\{Z_t\}$,\\[0.3mm]
         \hspace*{12.6mm} transfer functions $G_1$, $G_2$, $H_1$ and $H_2$ parame-\\[0.3mm]
         \hspace*{12.6mm} terized by $\theta \in \Theta$ and $\gamma \in \Gamma$,\\[0.3mm]
         \hspace*{12.6mm} user-chosen significance level $\alpha \in (0,1)$,\\[0.3mm]
         \hspace*{12.6mm} confidence sets $\widehat{\Theta}$ and $\widehat{\Gamma}$ for $\theta_*$ and $\gamma_*$
         respec-\\[0.3mm]
         \hspace*{12.6mm} tively with confidence level at least $1-\beta$,\\[0.3mm]
         \hspace*{12.6mm} dependence measure estimator $\widehat{\Delta}$,\\[0.3mm]
         \hspace*{12.6mm} tie breaking permutation $\sigma$ on $[m-1]_{\scriptscriptstyle 0}$\\[-2mm]
    \begin{algorithmic}[1]
         \hrule
         \STATE Choose integers $1 \leq r \leq m$ such that $\nicefrac{r}{m}  \leq \alpha - 2\beta$.\\[1mm]
         \STATE Generate $m-1$ random permutation $\{\pi_j\}_{j=1}^{m-1}$ \\[1mm]
         uniformly from $S_n$.\\[1mm]
         \STATE Construct noise term functions for $t=1, \dots, n$ by \\[1.5mm]
         $E_t(\theta) \doteq H_1^{-1} (q^{-1}, \theta)\big( Y_t -  G_1(q^{-1}, \theta) U_t \big)$,\\[1.5mm]
         $N_t(\gamma) \doteq H_2^{-1} (q^{-1}, \gamma)\big( Z_t -  G_2(q^{-1}, \gamma) V_t \big)$.\\[1.5mm]
         \STATE Build $m-1$ new alternative sample functions by\\[1mm]
         $\CD_j(\theta,\gamma)\hspace{-0.6mm} = \hspace{-0.6mm}\{(E_i(\theta),N_{\pi_j(i)}(\gamma))\}_{i=1}^n$ for $j \in [m-1]$\\[1mm]
         and let $\CD_j^\sigma(\theta,\gamma)= (\CD_j(\theta,\gamma),\sigma(j))$ for $j\in[m-1]_{\scriptscriptstyle 0}$.
         \STATE Formulate the dependence measure estimates\\[1mm]
         $\widehat{\Delta}_n^{(j)}(\theta, \gamma) = |\widehat{\Delta}(\CD_j(\theta, \gamma))|$ for $j = 0, 1, \dots, m-1$.\\[1mm]
         \STATE Construct the parameter-dependent ranking function
         \begin{equation*}
         \begin{aligned}
             & \psi_\Delta(\theta, \gamma) \doteq \psi_\Delta\big( \CD_0^\sigma(\theta, \gamma), \dots, \CD_{m-1}^\sigma(\theta, \gamma)\big)\\[1mm]
             & = 1 + \sum_{j=1}^{m-1} \BI\,\big(\,\widehat{\Delta}_n^{(0)}(\theta, \gamma) \prec_\sigma \widehat{\Delta}_n^{(j)}(\theta, \gamma)\,\big).
         \end{aligned}
         \end{equation*}
         \STATE Reject the null hypothesis if and only if\vspace{-1mm}
         \[\max_{\theta \in \widehat{\Theta}, \gamma \in \widehat{\Gamma}}\psi_\Delta( \theta, \gamma ) \leq r.\]\\[1mm]
    \end{algorithmic}
    \vspace*{-2mm}
\end{algorithm}

\subsection{Strong Consistency of the Robust Test}
In this section we quantify the asymptotic behaviour of rejection probability when $H_1$ is true, thus let us consider the case when $E$ and $N$ are dependent. We prove that the power of the suggested test tends to $1$ as the sample size goes to infinity. 
Let $B(\theta, \varepsilon)$ denote the Euclidean ball around $\theta$ with radius $\varepsilon$. We assume that:
\begin{itemize}
        \item[A\theasscounter] {\em Control inputs $\{U_t\}$, $\{V_t\}$ and driving noises $\{E_t\}$, $\{N_t\}$ are a.s.\ included in a C{\'e}saro space for $p=\infty$, i.e., for $\{W_t\} \in \{\{U_t\},\{V_t\},\{E_t\},\{N_t\}\}$ we have\vspace*{-1mm}
        \[\norm{W}_{c(\infty)} \,\doteq\, \sup_{n \in \NN} \frac{1}{n}\sum_{t=1}^n |W_t| \,<\,\infty. \vspace*{-1mm}\]}
        \addtocounter{asscounter}{1}
        \item[A\theasscounter] {\em There a.s.\ exist $K, \tilde{\varepsilon} > 0$ such that for $\theta \in B(\theta_*, \tilde{\varepsilon})$:
        \[ \norm{E(\theta_*) - E(\theta)}_{c(\infty)}\, \leq\, K \cdot \norm{\theta_* - \theta},\]
        and respectively for $N(\gamma)$, where $\gamma \in B(\gamma^*, \tilde{\varepsilon})$.}
        \vspace{2mm}
        \addtocounter{asscounter}{1}
        \item[A\theasscounter] {\em The confidence sets are uniformly consistent, i.e., for all $\varepsilon > 0$ there a.s.\ exists an $N_0 \in \NN$ such that for all $n > N_0$ both $\widehat{\Theta}_n \subseteq B(\theta_*, \varepsilon)$ and $\widehat{\Gamma}_n \subseteq B(\gamma_*, \varepsilon)$.}
        \addtocounter{asscounter}{1}
        \vspace{2mm}        
        \item[A\theasscounter] {\em Dependence measure estimator $\widehat{\Delta}$ is Lipschitz continuous around $(\theta_*, \gamma_*)$, i.e., $\exists \,C, \dbtilde{\varepsilon} > 0$ such that 
        \[ \hspace*{-3.3cm}|\widehat{\Delta}\big(\CD_j(\theta_*, \gamma_*)\big) - \widehat{\Delta}\big(\CD_j(\theta, \gamma)\big)| \]
        \[\leq C \cdot \Big(\norm{E(\theta_*)- E(\theta)}_{c(\infty)} + \norm{N(\gamma_*)- N(\gamma)}_{c(\infty)}\Big)\]
        for $\theta \in B(\theta_*, \dbtilde{\varepsilon}\,), \gamma \in B(\gamma_*, \dbtilde{\varepsilon}\,)$ and $j =0, \dots, m-1$.}
        \addtocounter{asscounter}{1}
\end{itemize}
{\em
\vspace*{-6mm}
\hspace*{-1mm}
\begin{theorem}
Assume $A1$ -- $A11$. If $H_1$ holds true, then
 \begin{equation}\label{eq:prob-rank_robust}
        \PP\Big( \max_{(\theta, \gamma) \in \widehat{\Theta}_n \times \widehat{\Gamma}_n} \psi_\Delta(\theta, \gamma) \leq r \,\Big) \xrightarrow{\,n\to \infty\,} 1.
    \end{equation}
\end{theorem}
}
\medskip
\begin{proof}
We use a characteristic $\Delta$, thus under $H_1$ we have
$\widehat{\Delta}_n^{(0)}(\theta_*, \gamma_*) \xrightarrow{p}\kappa\doteq |\Delta(E,N)| > 0$. In addition, $\widehat{\Delta}_n^{(j)}(\theta_*, \gamma_*) \xrightarrow{p} 0$ for $j \in [m-1]$
because of $A7$. Let us fix a positive $\varepsilon$ that is smaller than $\tilde{\varepsilon}$ and $\dbtilde{\varepsilon}$. By $A10$, there exists almost surely an $N_0 \in \NN$ such that $\widehat{\Theta}_n \in B(\theta_*, \varepsilon)$ and $\widehat{\Gamma}_n \in B(\gamma_*, \varepsilon)$ for all $n > N_0$. We prove that for $n$ large enough the rank values equal to $1$ uniformly on $B(\theta_*, \varepsilon) \times B(\gamma_*,\varepsilon)$ with large probability. We know that $\widehat{\Delta}_n^{(j)}(\theta_*, \gamma_*)$ is closer than $\varepsilon$ to $\kappa$ with large probability if $n$ is large enough.
Then, condition $A9$ and $A10$ yield
\begin{equation*}
\begin{aligned}
    &|\widehat{\Delta}_n^{(0)}(\theta_*, \gamma_*) -  \widehat{\Delta}_n^{(0)}(\theta, \gamma)|\\
    &\leq C \cdot \big(\norm{E(\theta_*)- E(\theta)}_{c(\infty)} + \norm{N(\gamma_*)- N(\gamma)}_{c(\infty)}\big)\\
    &\leq C \cdot K \cdot \big( \norm{\theta_* -\theta} + \norm{\gamma_* - \gamma}\big),
\end{aligned}
\end{equation*}
which proves that there exists $C_1 > 0$ such that a.s. 
\begin{equation*}
\begin{aligned}
    \sup_{\theta \in B(\theta_*, \varepsilon), \gamma \in B(\gamma_*, \varepsilon)}\big|\widehat{\Delta}_n^{(0)}(\theta_*, \gamma_*) -  \widehat{\Delta}_n^{(0)}(\theta, \gamma)\big| \leq C_1 \cdot \varepsilon.
\end{aligned}
\vspace{-2mm}
\end{equation*}
Thus, the infimum of $\widehat{\Delta}_n^{(0)}(\theta, \gamma)$ on confidence set $\widehat{\Theta}_n \times \widehat{\Gamma}_n$ tends to $\kappa$ in probability.
Similarly one can prove that $\sup\widehat{\Delta}_n^{(j)}(\theta, \gamma)$ on $\widehat{\Theta}_n \times \widehat{\Gamma}_n$ tends to $0$ for $j=1, \dots, m-1$, implying that the probability of rejection goes to $1$.
\end{proof}

\section{Illustrative Example: AR(1) Systems}

We considered two AR(1) systems: $Y_t=\alpha_* Y_{t-1} + E_t$, and $Z_t= \beta_* Z_{t-1} + N_t$ with $\alpha_*= 0.5$ and $\beta_* = 0.3$.
For parameters $(\alpha,\beta)$ the residuals $\{E_t(\alpha)\}$ and $\{N_t(\beta)\}$ can be computed from $\{Y_t\}, \{Z_t\}$ as
$E_t(\alpha) = Y_t - \alpha Y_{t-1}$
and similarly for $\{N_t(\beta)\}$.
For AR(1) systems, the required Lipschitz continuity 
can be easily satisfied. If $|\alpha| < 1$, then under $A8$  one can prove $\norm{Y}_{c(\infty)} \!< \infty$ and $A9$. For HSIC estimates, presented in \eqref{eq:hsic}, $A7$ holds and if $k\otimes \ell$ is Lipschitz and characteristic, then $A11$ is also satisfied.  

\begin{figure*}[t!]
    \centering
	\hspace*{-1mm}	
 	\subfigure[HSIC]{\label{fig:ref_vals_hsic}\includegraphics[height=41mm, width = 41mm]{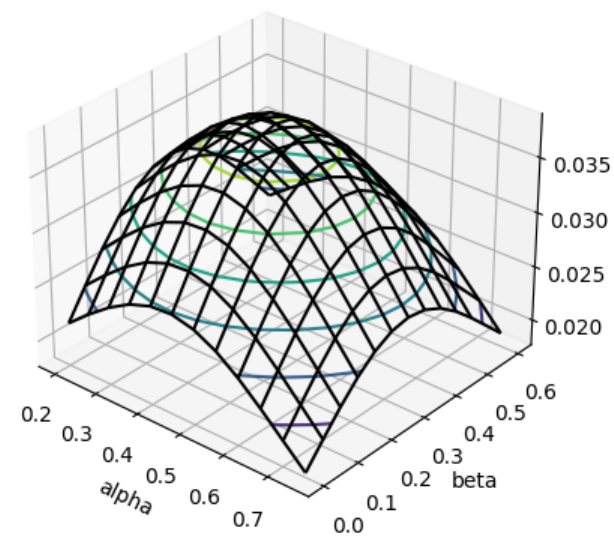} }
        \subfigure[Ranks (HSIC)]{\label{fig:ranks_hsic}\includegraphics[height=41mm, width = 41mm]{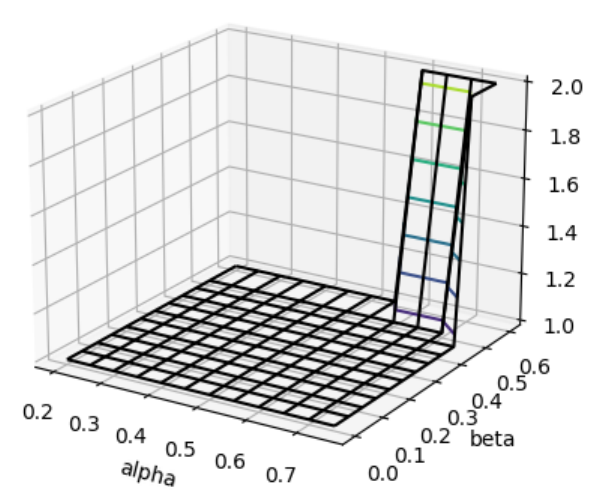}}
        \subfigure[Distance Covariance]{\label{fig:ref_vals_dist}\includegraphics[height=41mm, width = 41mm]{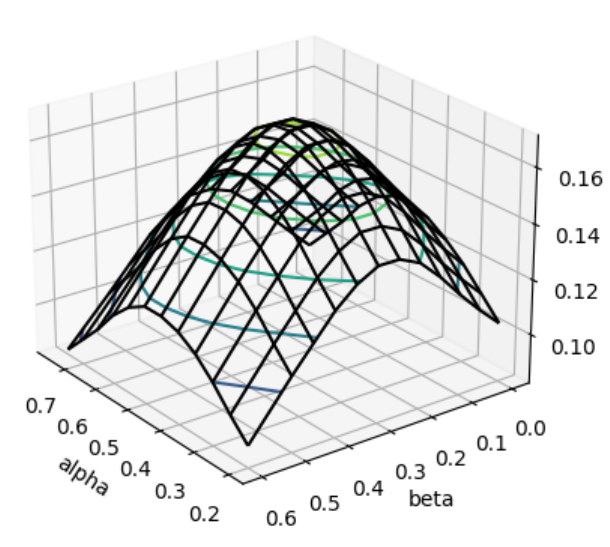}} 
        \subfigure[Ranks (Dist.~Cov.)]{\label{fig:ranks_dist}\includegraphics[height=41mm, width = 41mm]{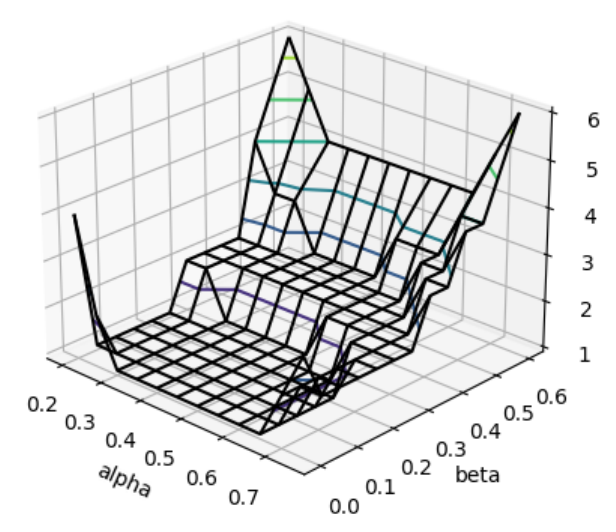}}
        \subfigure[Rotated Gaussian Mixture]{\label{fig:rot}\includegraphics[height=41mm, width = 41mm]{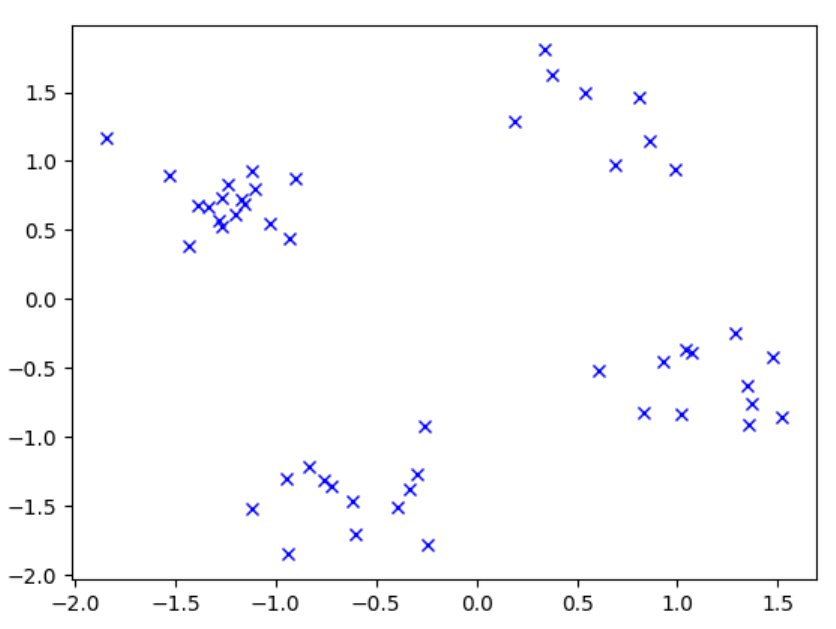}}
        \subfigure[Power Functions]{\label{fig:power_rot}\includegraphics[height=41mm, width = 41mm]{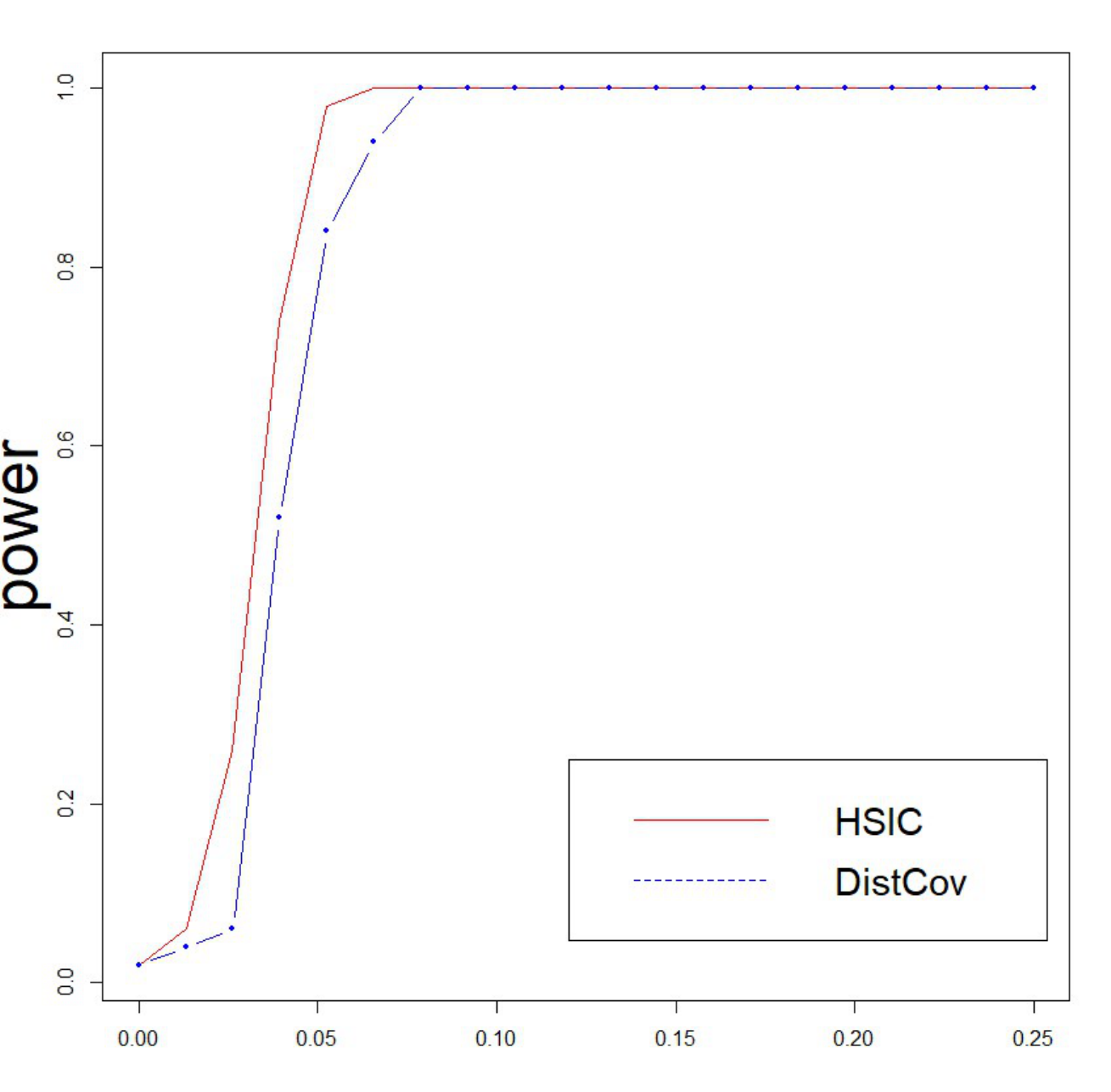}}
	\subfigure[Extinct Gaussian]{\label{fig:ext}\includegraphics[height=41mm, width = 41mm]{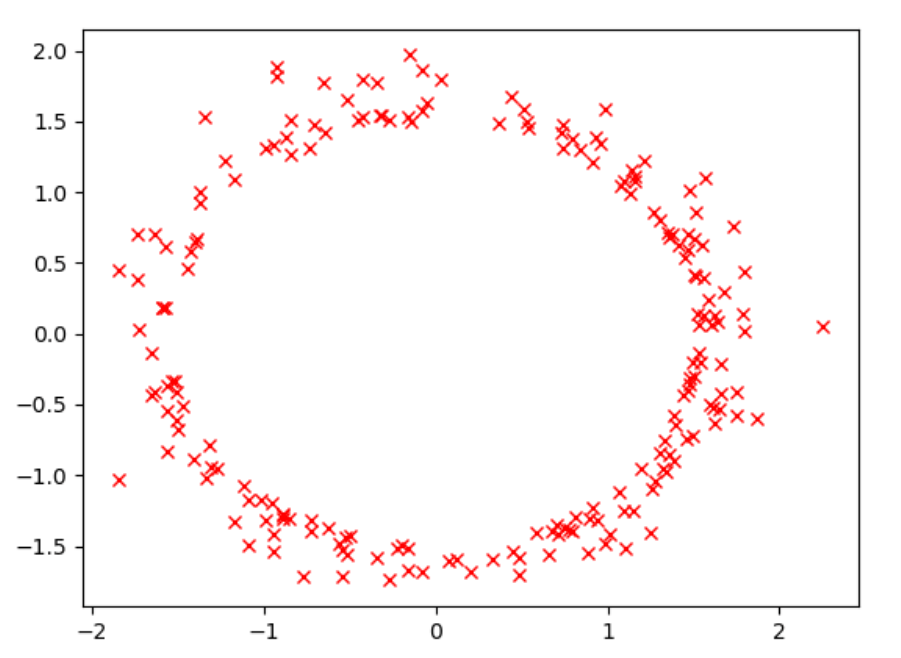}}
	\subfigure[Power Functions]{\label{fig:power_ext}\includegraphics[height=41mm, width = 41mm]{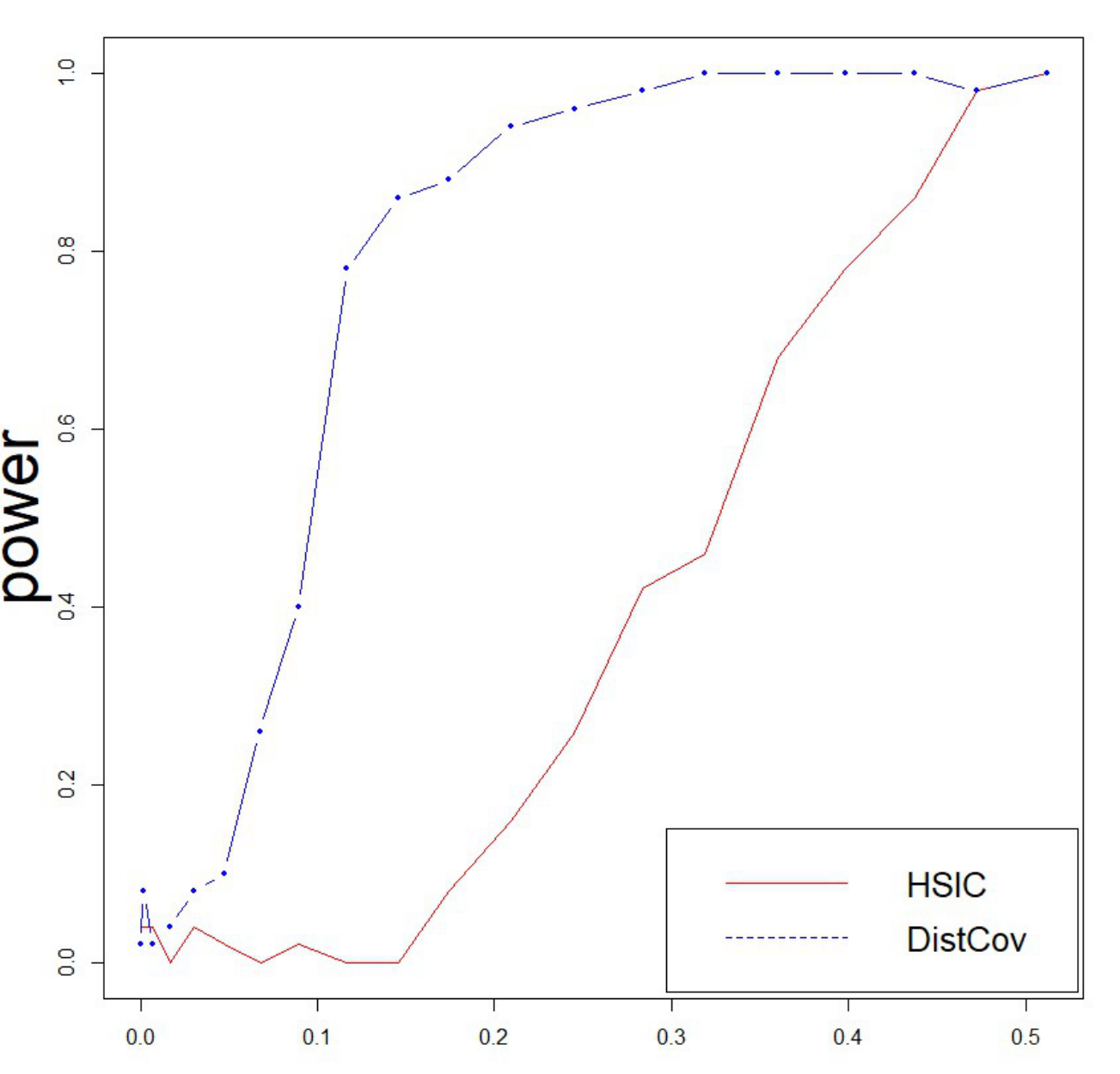}}
    \vspace{-1mm}
    \caption{Reference functions, ranking functions, innovation data and power functions for AR(1) systems.}
\label{fig:experiments}
\vspace*{-3.5mm}
\end{figure*} 

\subsection{Sign-Perturbed Sums}
For ARX systems, confidence sets satisfying $A10$ can be constructed, e.g., by the Sign-Perturbed Sums (SPS) method \cite{csaji2014sign}. 
{ Standard SPS assumes independent and symmetric noises.
Under the i.i.d.\ assumption, symmetry is no more required \cite{kolumban2015perturbed}.}
An instrumental variable-based extension of SPS was proposed in \cite{volpe2015sign}, which is uniformly consistent, even for ARX systems with feedback control.
\subsection{Numerical Simulations}
{ We simulated two AR(1) systems with nonlinearly dependent
i.i.d.\ 
innovations $\{(E_t,N_t)\}_{i=1}^n$. The 
processes were initialized in zero ($A4$).} First, a rotated mixture of Gaussian distributions \cite{gretton2007kernel} was considered for innovations, see Figure \ref{fig:rot}. We generated a sample with $n=50$ elements from a zero mean two-dimensional Gaussian distribution
with covariance matrix $\nicefrac{1}{4}\cdot\BI$, then we shifted each data-point with a pair of random signs and rotated the obtained points around the origin with an angle of $0.1$ (radian). We used SPS to construct confidence sets for $\alpha_*$ and $\beta_*$ with significance level $\nicefrac{1}{80}$ and tested independence with $m=40$ datasets. We maximized the ranking function on a fine grid of the confidence region.

Reference 
functions $\widehat{\Delta}_n^{(0)}$ and ranking functions $\psi_\Delta$ are plotted on Figure \ref{fig:ref_vals_hsic}, \ref{fig:ref_vals_dist}, \ref{fig:ranks_hsic} and \ref{fig:ranks_dist} for HSIC and distance covariance. At significance level $0.15$ based on Figure \ref{fig:ranks_hsic} we reject the null hypothesis, but we accept $H_0$ at this level based on Figure \ref{fig:ranks_dist}, because $\psi_\Delta$ exceeds $5$ at some points.
The (estimated) power functions, i.e., the rejection probabilities,
are plotted for $n=200$ and significance level $0.15$ on Figure \ref{fig:power_rot} w.r.t.\ the rotation angle, which served as a factor inducing dependence. 

Second, a sample from an extinct 
multivariate Gaussian 
distribution with covariance $\nicefrac{1}{4}\cdot\BI$ was used to generate innovations, see Figure \ref{fig:ext}. That is, we introduced dependence between $E$ and $N$ by throwing away the pairs that lied in a circle around the origin with radius $r$. Figure \ref{fig:power_ext} shows the {(estimated)} power functions w.r.t.\ the distinction rate increasing with $r$ for $n=500$ using HSIC and distance covariance with  significance level $0.15$.

\section{Conclusion}
In this paper we introduced hypothesis tests for the independence of synchronous 
general linear systems with non-asymptotically guaranteed significance levels.
The main idea was to apply permutation tests over a confidence region for the system parameters. We combined these ideas with characteristic dependence measures to detect any nonlinear dependence between the innovations of the systems.
We proved consistency under general assumptions and demonstrated the method on AR(1) systems.\!\!

%
                                  %
                                  %
                                  %
                                  %
                                  %

%

%

%
%

%

%

\vspace*{-1mm}
\bibliographystyle{ieeetr}
\bibliography{refs} 
\vspace*{-2mm}

\end{document}